\documentclass[letterpaper, 10 pt, conference]{ieeeconf}  

\IEEEoverridecommandlockouts                           

\usepackage{amsmath, amssymb , graphicx, subcaption}
\usepackage{multirow}
\usepackage{dblfloatfix} 
\usepackage{balance}
\usepackage{hyperref}

\DeclareCaptionFont{caption}{\fontsize{8}{9.6}\selectfont}
\title{\LARGE \bf
An Open-Source 7-Axis, Robotic Platform to Enable Dexterous Procedures within CT Scanners}

\author{Dimitri A. Schreiber$^\dag$, Daniel B. Shak$^\dag$, Alexander M. Norbash$^\ddag$, and Michael C. Yip$^\dag$, \IEEEmembership{Member, IEEE}
\thanks{$^\dag$Department of Electrical and Computer Engineering, University of California San Diego, La Jolla, CA 92093 USA. {\tt\small \{dschreib, dbshak, m1yip\}@eng.ucsd.edu}
$^\ddag$Department of Radiology, University of California San Diego, La Jolla, CA 92093 USA. {\tt\small anorbash@ucsd.edu}}}

\begin{document}

\maketitle
\thispagestyle{empty}
\pagestyle{empty}

\begin{abstract}
This paper describes the design, manufacture, and performance of a highly dexterous, low-profile, 7 Degree-of-Freedom (DoF) robotic arm for CT-guided percutaneous needle biopsy. 
Direct CT guidance allows physicians to localize tumours quickly; however, needle insertion is still performed by hand. This system is mounted to a fully active motion stage superior to the patient's head and teleoperated by a radiologist. Unlike other similar robots, this robot's fully serial-link approach uses a unique combination of belt and cable drives for high-transparency and minimal-backlash, allowing for an expansive working area and numerous approach angles to targets, all while maintaining a small in-bore cross-section of less than $16cm^2$. 
Simulations verified the system's expansive collision free work-space and ability to hit targets across the entire chest, as required for lung cancer biopsy. Targeting error was on average $<1mm$ on a teleoperated accuracy task, illustrating the system's sufficient accuracy to perform biopsy procedures. 
While this system was developed for lung biopsies, it can be easily modified for other CT-guided needle based procedures. 
Additionally, this system is open-hardware.\footnote{\label{file_link}Files are available at: \url{https://github.com/ucsdarclab/Open-Source-CT-Biopsy-Robot}}
\end{abstract}

\section{INTRODUCTION}
Primary lung cancer is by far the leading cause of cancer death worldwide. When symptoms arise, the five-year lung cancer survival rate is 17\% \cite{cancerorg}.  Early lung cancer detection through screening with low-dose CT and needle biopsy has been shown to reduce mortality for high-risk patients. Definitive diagnosis of lung cancer requires tissue sampling, often performed by percutaneous transthoracic CT-guided lung biopsy. Smaller lesions, especially those that are under 1cm in size, are challenging to target with current hand-guided methods\cite{Tian2017}. 

During a needle lung biopsy, the radiologist must localize and target a pulmonary lesion several centimeters below the surface of the skin. The radiologist moves a patient in and out of a CT scanner when alternating between viewing the position so f the needle and nodule on the CT scan in a separate room and manually advancing the needle. The non-real-time 3D scans and the freehand adjustments often result in the need for multiple punctures \cite{heerink2017complication}, and restrict the angles in which the needle can be stepped with confidence. This limits the radiologist's ability to reach lesions that require an atypical approach to avoid sensitive vessels or bone. Furthermore, the repetitive CT scans have significant risk of secondary imaging-induced cancer. Robotics can serve as a technological aid to significantly reduce these risks \cite{kettenbach2015robotic}.

\begin{figure}[tb]
    \centering
      \includegraphics[width=\linewidth,trim={0 0 10cm 0},clip=true]{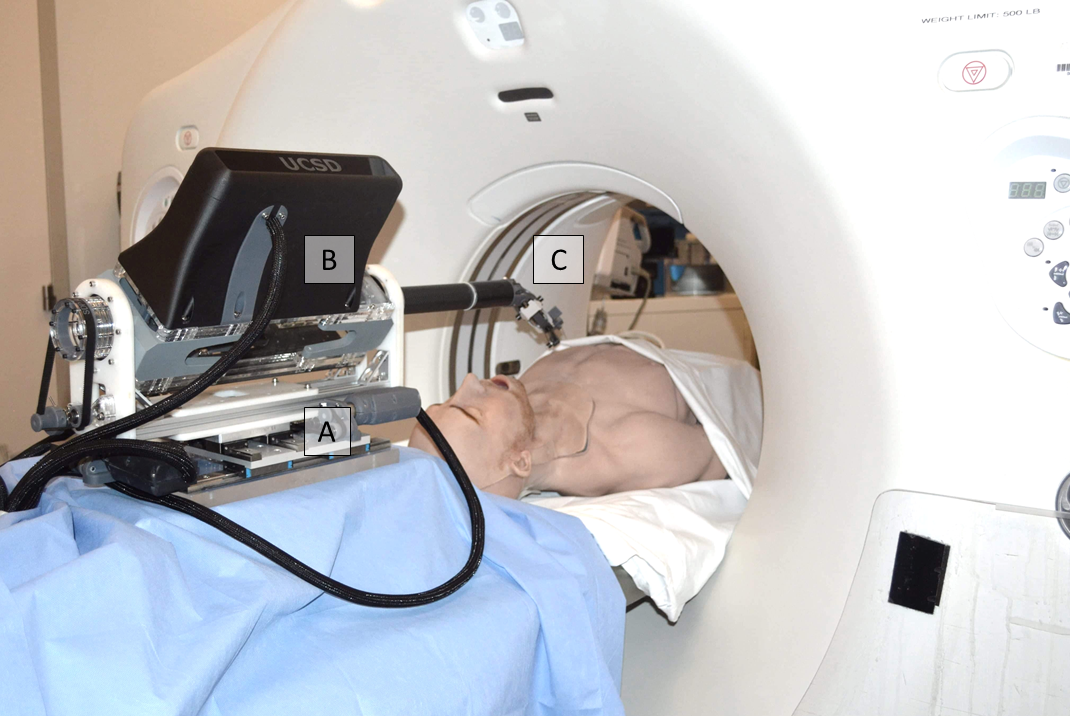}
    \caption{CT-guided robotic system with a low-intra-bore-profile serial-link redundant 7-DoF arm. \boxed{A} is the cartesian positioning stage. \boxed{B} is the trunnion and rotary axis. \boxed{C} is the robote 4DoF arm.}
    \label{fig:robot_person_UCSD}
\end{figure}

Here, we present a teleoperated robot for intra-bore CT-guided lung biopsies (Fig. \ref{fig:robot_person_UCSD}). The robot uses a redundant serial linkage design that offers 7-DoF cable-driven control for minimal-backlash and smooth positioning and inserting of biopsy needles. This robot offers significant increases in the number of approach vectors to peripheral targets. The robot reaches into the entire length of the bore and can maneuver through it in a wide sweeping positional and orientational working volume. End-to-end development from mechanical design to full user interface was required to achieve both full dexterity but also high-transparency and intuitive teleoperation. \\
Thus, we provide the following set of contributions:
\begin{enumerate}
    \item A highly dexterous 7-DoF robotic biospy arm leveraging serial linkage design with outlined CT-compatible design considerations,
    \item a zero-backlash, stiff transmission through the use of belt and cable drives as required for high-fidelity teleoperation,
    \item an intuitive teleoperational interface and control for direct image guided intervention while the patient remains within the scanner bore,
    \item complete simulation environment in V-REP with full workspace analysis demonstrating reach, and
    \item a fully open-source approach to engage the robotics community.
\end{enumerate}

We present description and analysis of the robot's mechanical design, characteristics, and capabilities. Furthermore, we verify the collision-free work-space through simulation, and perform experiments to determine the system's repeatability, teleoperated accuracy. In addition, a phantom biopsy procedure is performed under CT guidance.

The proposed robot will enable a radiologist to quickly target a lung nodule, precisely adjust and align the needle's trajectory with the nodule, and biopsy the nodule. This will reduce the number of punctures and CT scans required for the procedure. However, given the large workspace and effective teleoperation of the robot, we furthermore see a potential to use this platform for a variety of different intra-bore applications. Thus, by open-sourcing the platform, we aim to contribute a base platform solution to the surgical robotics community. Our design overcomes many coupled mechanical, electrical, user-interface, and systems engineering challenges. This reduces the barrier for other researchers in surgical robotics working toward 3D image-guided teleoperation or automation research.

\section{Related Work}

Robotic systems have been developed to address a range of image-guided needle biopsy applications ranging from liver ablation to prostate brachytherapy \cite{kettenbach2015robotic,number46}. Yet only a select few have been designed to address the specific challenges of performing lung biopsy \cite{solomon2002robotically,anzidei2015preliminary,kuntz2016toward,ben2018evaluation, dou2017design,moon2015development,Zhou2011}. 

Prior robotic approaches to lung biopsy can be broadly categorized by their physical approach to lesions, the number of active joints, their mechanical stiffness, and their controllers. Both platforms that use industrial robot arms and those with passive setup joints for transthoracic biopsy have a reduced number of feasible biopsy approach due to their limited reach into the bore due to their arm size \cite{Atashzar2013,Zhou2011,Anzidei2015} and  limited active range of motion, respectively. They may be well suited for smaller anatomies with single approach vectors. Robotic systems using a device secured to the chest \cite{walsh2007evaluation,Seitel2009,Hungr2016} have limited reach but may have better precision for positioning instruments as the body shifts. 

Other robotic systems \cite{swaney2017toward,kratchman2011toward,van2015design,swaney2015tendons,reed2011robot}, including recently released commercial lung biopsy systems \cite{ion,auris}, offer a bronchoscopic approach via intraluminal steerable needles \cite{rojas2018robotic,fielding2017first}. However, they have low diagnostic yield in peripheral lungs and for lesions smaller than 20mm \cite{memoli2012meta,gilbert2014novel}, \cite{han2018diagnosis,aerts2016transthoracic}. 

The success of these application-specific CT compatible robotics systems illustrates the utility of CT-guided needle biopsy robots. However, due to the tight working conditions combined with stringent material limitations, existing robots have deficiencies in their operating area and reachable collision-free workspace. This precludes them from general purpose application. The proposed serial linkage robotic system overcomes several of these challenges by striking a balance between the high-stiffness industrial robot arm methods and the low-stiffness intra-bronchoscopic needle approaches. 

\section{Methods}
This section describes the clinical requirements for lung biopsy and the considerations required for teleoperated robot design. This includes the mechanical design, the electrical interfaces, kinematics, and user interface for teleoperation.

\subsection{Clinical Requirements}
Clinical requirements for lung biopsy includes the ability to hit 10mm biopsy targets with an average needle placement depth of 73mm (4.8-139.6 range)\cite{number15}. Commonly used needles range in length from 8cm to 30cm. The average human lungs are 25cm to 35cm in length and 10 to 15cm in width. Vasculature and ribs preclude certain approaches and other approaches are sub-optimal due to their high risk of heart and diaphragm puncture. This motivates a highly dexterous robot with a large working volume to enable access to target nodules from several orientations throughout the lungs without reconfiguration and manual setup. The robot must be radiolucent to minimize visual artifacts in the CT images.

Teleoperation enables direct ``surgeon in-the-loop control''. Humans naturally adjust for bias resulting from kinematic model inconsistencies with the real robot, as evidenced from the well-accepted teleoperation of the daVinci Surgical System for laparoscopic surgery (the robot uses a serial linkage approach). Furthermore, due to the radiologist's extensive training, they can compensate for needle deflections which occur due to tissue non-homogeneity and small initial placement errors. This places smooth and fine adjustability as a higher priority than pure kinematic accuracy and stiffness. 




\subsection{Linkage Design}

Kinematically and mechanically, the robot is composed of two distinct structures. The 3-DoF exo-bore motion stage is a linear-linear-roll belt driven positioning platform. Each actuator for this platform resides on the previous link. The 4-DoF cable driven in-bore actuator is kinematically a yaw-pitch-yaw-translational assembly. The four actuators for this assembly reside on a trunnion which serves as the structure for the system's rotary axis.

For the serial cable driven joints, the number of idler pulleys increases with order $N^2$, where $N$ is the number of joints. Therefore, it is beneficial to limit the number of serial cable driven joints required. This platform's 4-DoF cable driven intra-bore arm with the 3-DoF positioning stage provides a compromise between design complexity and system dexterity. This maximizes the platform's in-bore performance while minimizing its cross-sectional area. Mechanical benefits to using the serial cable driven design include:
\begin{itemize}
\item  \textit{Zero Backlash:} The use of cables with pulleys (in contrast with Bowden tubes) allows the arm to have zero backlash and high transparency which provides fine incremental motion. Additionally, there is no ``stiction'' which would interfere with fine adjustment. 
\item \textit{High Load Bearing:} In contrast with belts, synthetic cables have high load bearing capabilities while still allowing flexible 3D routing and tolerating a small pulley diameter to cable diameter bending radius (D:d).
\item \textit{Improved Dexterity and Volume:} The serial link design contributes to the robot's high dexterity while minimizing its in-bore profile. Motors are placed out of the bore, where space is less constrained.
\item \textit{Reduced Mass:} The remote drive significantly decreases the arm's static loading, as each joint does not have to hold the following joints' actuators.
\item \textit{Improved Image Quality:} There is a significant reduction in imaging artifacts in comparison with joint mounted actuators.
\end{itemize}

For needle placement, the combined 7-DoF design (including 1 needle advancement access) provides two redundant DoF (a needle is symmetric around its roll axis) which can be used for external collision avoidance via nullspace control.

A long carbon-fiber tube with the terminal 4-DoF cable driven end-effector reaches into the bore to place the dexterous end in the general working area. Carbon-fiber is both radiolucent (due to its low density) and possesses very high stiffness. The out-of-bore stage's two prismatic axes enable motion within the frontal plane. The revolute axis enables the trunnion to roll in the transverse plane. 

Linear ball bearing rails with a belt drive are used for the stage's prismatic axis due to their high stiffness, accuracy, and low friction, and lack of backlash, respectively. Belt drives provide an excellent solution for backlash-free motion when space is less constrained, and the routing path is not tortuous. This decoupling of the intra-bore and exo-bore components enables our 7-DoF arm to have less than a 50mm$\times$50mm frontal cross section in the bore, increasing the use-able working space and decreasing the impingement upon the patient's area. Furthermore, image artifacts are virtually eliminated (see Fig. \ref{fig:CT_Biopsy_puncture}) due to the few dense materials present in the bore. The design has 0.3m travels in both the robot's x-axis and y-axis.

Maxon RE339152 (24V, 10900 free RPM) motors connected to Maxon 370805 planetary gearbox (3.4Nm rated torque, 479:1 ratio) with US Digital E4T 500 count encoders on the motor shaft are used for all axis. This results in $3.7e-4$ degrees per count encoder resolution at the output shaft.

The robot end-effector is manufactured using a combination of Selective Light Activation 3D printing, continuous fiber reinforced Fused Deposition Modeling 3D printing, laser-cut acrylic plates, and turned pulleys. The robot base is constructed of plastics (Delrin, PTFE, Acrylic-like materials, nylon), composites (carbon and glass fiber composites), and minimal metallic components (steel bearings, rods, bolts).

\subsection{Linkage Analysis}

The up-to-10N loads applied when inserting and sampling a tissue during a lung biopsy result in elastic deformation of the robot's remote 4-DoF joint and cables, and may cause brinelling of the idler pulleys' bearings. Tested lung biopsy forces are measured below 4N, but, they may be as high as 8N\cite{259616}. Assuming needle loading primarily acts as thrust load and torques applied to the needle, and consequently the insertion joint, are minor, the maximum joint torques for the three revolute joints on the 4-DoF arm can be bounded by:
\begin{align}
    &\tau_{q_4} = \max_{x}[sin(x)(0.16\text{m})(8 \text{N})]\nonumber\\
    &\tau_{q_5} = \max_{y}[sin(y)(0.08\text{m})(8)\text{N})]\\
    &\tau_{q_6} = 0 \nonumber
\end{align}
Therefore, the maximum applied load from the needle biopsy is 1.28Nm, 0.64Nm, and 0Nm for joints 4, 5, and 6 respectively.

The 4-DoF cable driven arm's idler pulley bearing load ratings primarily limits the robot's load tolerance. The R2-5 stainless steel bearings (McMaster Carr P/N 57155K347) have a static load rating of 177.8N. For the 4$^\text{th}$, 5$^\text{th}$, and 6$^\text{th}$ joints, this corresponds to maximum joint torques at the lowest load rating configuration of greater than 2.49Nm, 1.25Nm, and 1.25Nm, respectively. Pulley force increases at higher deviations from straight for joints 5, 6, and 7. Joint 7 has a bearing limited force rating of 177.8N. Therefore, the arm has a sufficient load rating to perform a core needle biopsy. The gravity torques ($\le 0.011$Nm) applied to joints of the cable driven arm are negligible compared to the torques due to needle insertion.

Cable stretch is, in general, a concern for precise control of serial linkage cable driven actuators. Disturbances due to cable stretch, unlike backlash, are linear. Nonetheless, it is still beneficial to minimize. The properties of several cable materials are summarized in Table \ref{tabl:cables}. We used Dyneema SK99 cable due to its ease of sourcing, high stiffness, low D:d ratio, acceptable creep, and UV resistance.

\begin{table}[h!]
    \centering
    \caption{\\Comparison of Cable Materials}
    \begin{tabular}{c|ccccc}
          Cable  & Tensile's  & Tensile  & D:d & Creep & Sourcing\\
          Type & Modulus & Strength & & Resistance & \\ \hline
          SK99 & 155 GPa & 4.1 GPa & 5:1 & Fair & Easy \\
          DM20 & 94 GPa & 3.4 GPa & 8:1 & Great & Difficult\\
          Vectran & 103 GPa & 3 GPa & 8:1 & Good & OK\\
          SS & 210 GPa & 2 GPa & 18:1 & Great & Easy \\
    \end{tabular}
    \label{tabl:cables}
\end{table}

\begin{figure}[b!]
    \centering
      \includegraphics[width=\linewidth]{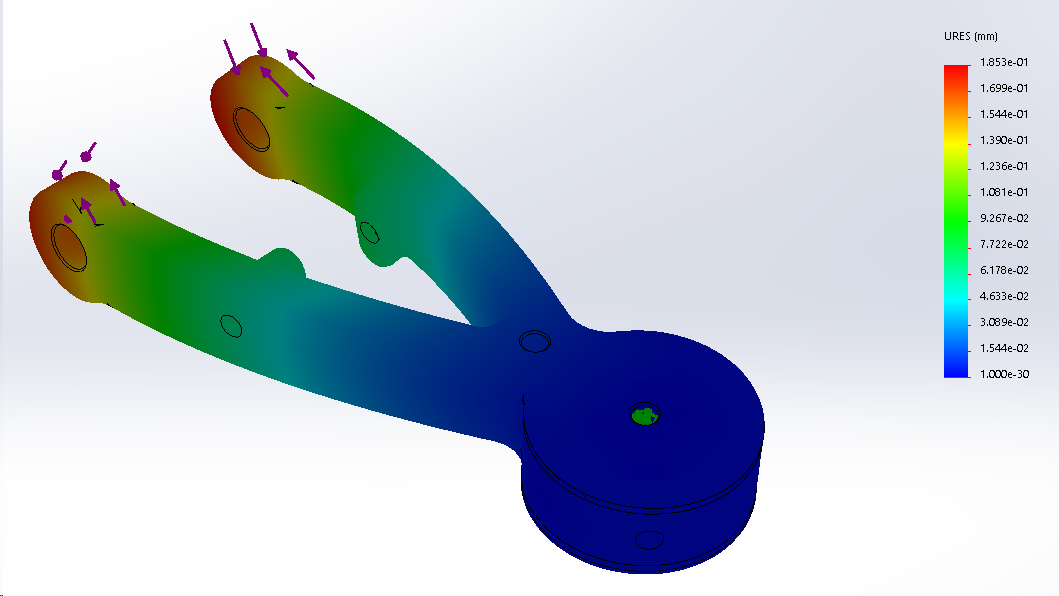}
    \caption{Finite Element Analysis deflection simulation with 2Nm applied torque for a single (base) linkage of the multi-linkage biopsy arm that extends into the bore.}
    \label{fig:fea}
\end{figure}

The linkages themselves are made of 3D-printed carbon-fiber-reinforced plastic. Finite Element Analysis (FEA) simulation of the base joint (which observes the highest deflection) is presented in Fig. \ref{fig:fea} using MarkForged's carbon fiber material properties. The FEA shows that the link deflection is negligible in comparison to the cable deflection.

To determine the worse-case deflection scenario, we consider a maximal deflection when the force is applied perpendicular to the outstretched arm (joint 4 with joint 5 straight). Due to negligible link deflection, the stiffness for this configuration is calculated as follows:
\begin{equation}
    \Delta L = \frac{F L_0}{A E}\quad \text{ and }\quad \Delta \theta = \frac{\Delta L}{2 \pi r}
\end{equation}
where $\Delta L$, the change in cable length, is calculated using the definition of Young's Modulus, $L_0$ is the static length, $F$ is the force applied to the cable, $A$ is the cross-sectional area of the cable, and $E$ is the cable's Young's Modulus. Using a first order approximation for tip deflection motion, this results in an end-effector stiffness of at least 1.55N/mm, or at most 0.64mm of movement for each 1N of additional force.

\begin{figure}[t!]
    \centering
    \vspace{2mm}

      \includegraphics[width=\linewidth]{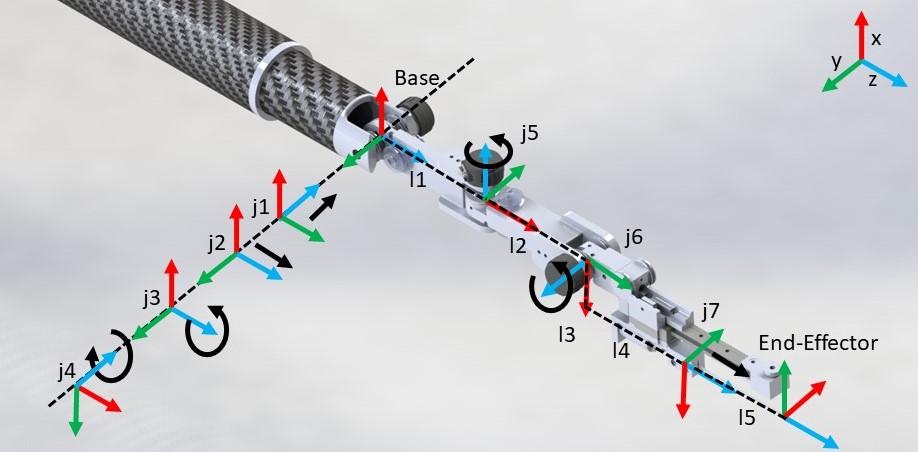}
    \caption{Kinematic diagram of the distal end of the robot, illustrating needle and base coordinate frames. A backend motion stage enables Cartesian positioning of the base joint.}
    \label{fig:kinematic}
\end{figure}



\subsection{Robot Kinematics}

The kinematic structure of the platform is shown in Fig. \ref{fig:kinematic}. The robot's kinematic chain is described using Modified Denavit-Hartenberg (DH) parameter. The DH parameters define the pose of the next frame by origin
\begin{equation}
    c_{n+1} = c_n + a x_n + d z_{n+1}
\end{equation}
and orientation
\begin{equation}
    R_{n+1} = 
    \setlength{\arraycolsep}{1pt}
    \renewcommand{\arraystretch}{0.6}
R_n \begin{bmatrix}
\centering
1  & 0  & 0 \\
0  & cos(\alpha) & -sin(\alpha)\\
0 & sin(\alpha) & cos(\alpha)  \\
\end{bmatrix}
\begin{bmatrix}
\centering
cos(\theta)  & -sin(\theta) & 0 \\
sin(\theta)  & cos(\theta) & 0\\
0 & 0 & 1  \\
\end{bmatrix}
\end{equation}
where $R_n$ is the n-th frame's orientation and $c_n$ is the n-th frame's position. For this robot, $n\subset \{1,...,7\}$ is the robot's reference frame for the joints and $ n = 8$ is the end-effector frame.

The modified DH parameters for this robot corresponding to the kinematic diagram presented in Fig. \ref{fig:kinematic} and are provided in Table \ref{tabl:dh_parameters}; the actuator-to-joint mixing matrix that converts unit steps in the actuator to unit steps in joint angles is presented in Table \ref{tabl:mixing_matrix}. In this table, $q_i$ and $m_i$ correspond to the $i$-th joint's configuration and actuator's configuration, respectively. The actuator-to-joint mixing matrix accounts for the pulley reductions of the 3-DoF stage's belt drives and joint coupling for 4-DoF cable-driven arm due to the varying amount of cable wrap as coupled joints move.

\begin{table}
    \setlength\tabcolsep{0.9em}
    \centering
    \caption{\\DH parameters where p is prismatic and r is revolute}
    \scriptsize
    \begin{tabular}{c|c|cccc}
          Frame & Type & a (meters)& $\alpha$ (rad) & D (meters) & $\theta$ (rad) \\ \hline
          1 & p & 0 & $\frac{\pi}{2}$ & $q_1$ & $0$ \\
          2 & p & 0 & $- \frac{\pi}{2}$ & $q_2$ & $0$ \\
          3 & r & 0 & 0 & 0 & $q_3$ \\
          4 & r & 0 & $\frac{\pi}{2}$ & 0 & $q_4 + \frac{\pi}{2}$ \\
          5 & r & 8e-2 & $\frac{\pi}{2}$ & 0 & $q_5$ \\
          6 & r & 8e-2 & $\frac{\pi}{2}$ & 0 & $q_6 - \frac{\pi}{2}$ \\
          7 & p & 5.57e-2 & $-\frac{\pi}{2}$ & $2.74e-2 + q_7$ & 0 \\
          8 & - & 0 & 0 & 1.15e-1 & $\frac{\pi}{2}$ \\

    \end{tabular}
    \label{tabl:dh_parameters}
\end{table}

\begin{table}
    \setlength\tabcolsep{0.4em}
    \centering
    \caption{\\Actuator-to-Joint mixing matrix}
    \scriptsize
    \begin{tabular}{c|c|ccccccc}
          Joint & Type & $m_1$ & $m_2$ & $m_3$ & $m_4$ & $m_5$ & $m_6$ & $m_7$ \\ \hline
          $q_1$ & p & 5.73e-3 & 0 & 0 & 0 & 0 & 0 & 0 \\
          $q_2$ & p & 0 & 5.73e-3 & 0 & 0 & 0 & 0 & 0 \\
          $q_3$ & r & 0 & 0 & 0.24 & 0 & 0 & 0 & 0 \\
          $q_4$ & r & 0 & 0 & 0 & 0.45 & 0 & 0 & 0 \\
          $q_5$ & r & 0 & 0 & 0 & -0.35 & 0.45 & 0 & 0 \\
          $q_6$ & r & 0 & 0 & 0 & 0.94 & -0.62 & 0.79 & 0 \\
          $q_7$ & p & 0 & 0 & 0 & -5.26e-3 & 3.23e-3 & -8.73e-3 & 6.35e-3 \\

    \end{tabular}
    \label{tabl:mixing_matrix}
\end{table}

\subsection{Embedded System Controller}
Low-level motor control is performed using synchronized Proportional Integral Derivative (PID) controllers on a DE0 Nano SOC FPGA development board with a custom motor control PCB for 8-axis brushed DC motor control (MAX14870 driver IC) with current sensing (INA169 amplifier). The DE0 Nano SOC development board is a combination of a dual-core ARM Cortex A9 Hard Processor System (HPS) with a Cyclone V FPGA, interconnected through shared memory. Motors directly interface with the FPGA for encoder pulse counting, PWM generation, and current sensing. Additionally, a watchdog timer and emergency stop are implemented through the FPGA. High-frequency PID position motor control (1kHz loop rate) is implemented through the HPS and communicates with the FPGA using Direct Memory Access. The HPS hosts a TCP/IP web server to allow the remote master PC to update the motor position setpoints and disable/enable the robot. This physically separates the high-frequency, latency sensitive control from the low-frequency, high-level control.

\subsection{User Interface}

\begin{figure}[b]
\centering
      \includegraphics[width=\linewidth]{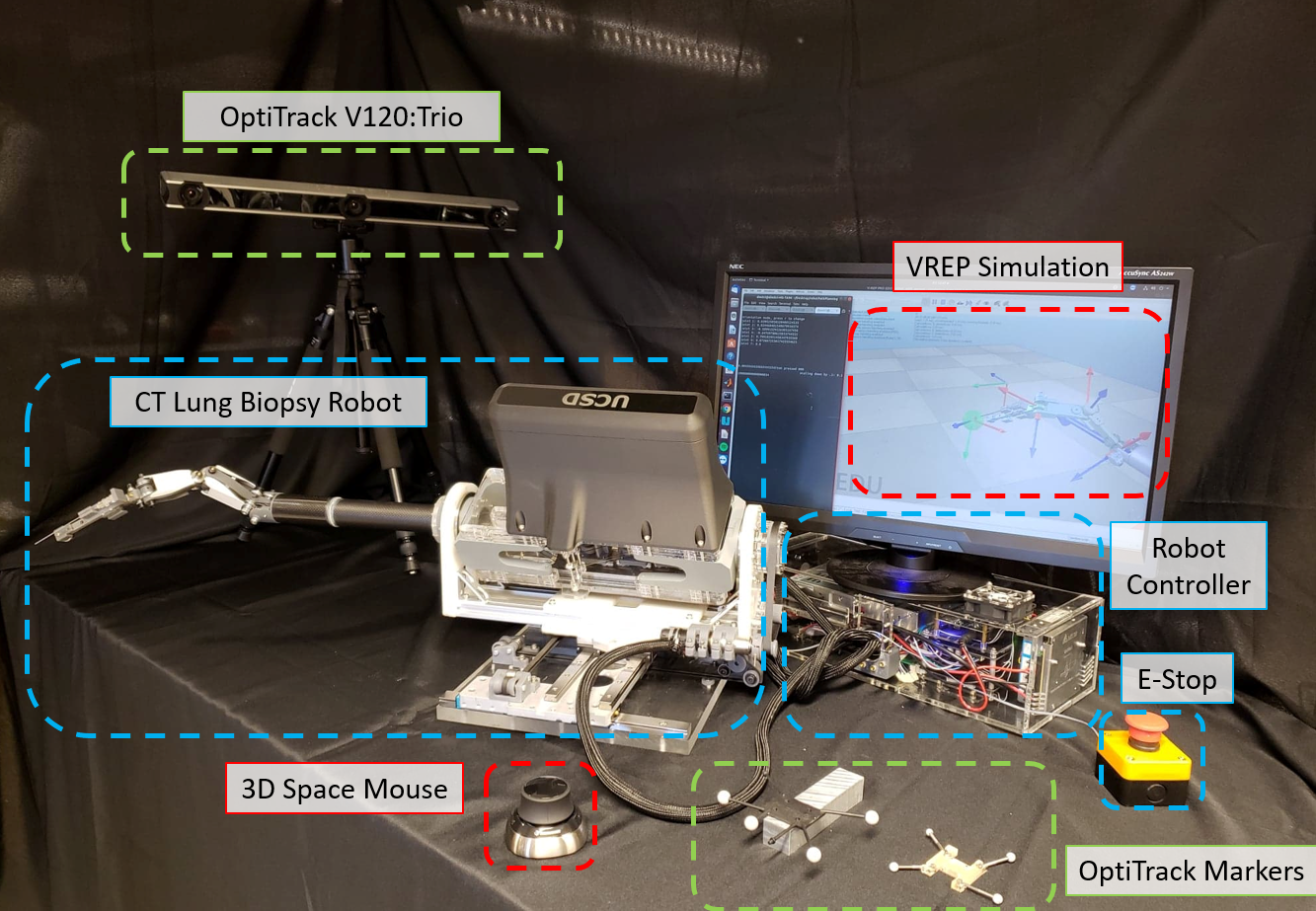}
    \caption{Experimental setup for the robot with Optitrack system and marker for benchtop tests (green box), robot and controller (blue box), and simulation with input device (red box).}
    \label{fig:optitrack}
\end{figure}

A kinematic simulation of the robot and collision simulation of the CT bore are developed using Coppelia Robotic's V-REP \cite{VREP2013}. A Python backend communicates with the simulator, interfaces with various peripheral devices, and brings the physical robot system together such that the simulator matches with the physical robot and the peripheral devices can be utilized to control either the simulation alone or the synchronized simulator and robot. The 3DConnexion SpaceMouse is used as the input device to update Frame 6's position $p[n]$ and rotation matrix $R[n]$ as follows:
\begin{align}
    p[n+1]&= p[n] + \gamma v[n]\\
    R[n+1]&= \textrm{euler2mat}(\gamma\cdot r[n]) R[n].
\end{align}
$\gamma \in (0,1]$ is a velocity scaling constant which can be dynamically adjusted using two auxiliary buttons on the 3DConnexion mouse, $v[n] \in \mathbb{R}^3$ is a discretized linear force reading from the 3DConnexion mouse in the $[x, y, z]^T$ axis, and $\gamma\cdot r[n] \in \mathbb{R}^3$ is an $\gamma$-scaled and discretized rotational velocity reading from the 3DConnexion mouse around $[roll, pitch, yaw]^T$ axis and converted into a rotation matrix via $\textrm{euler2mat}(\cdot)$. Inverse kinematics is calculated using a damped-least-squares method, available through V-REP, to solve for the joint configuration. This discrete time system is sampled at 400Hz. The final joint, needle insertion, is advanced and retracted using the control computer's keyboard in 0.1mm increments. 

For visual feedback, the radiologist who is teleoperating the robot, views the CT images on the computer present in the standard imaging suite for CT scanners. Additionally, a 1.6mm diameter video endoscope (Misumi MD-V1001L-120) is mounted parallel to the needle, and is displayed to the user to provide a first-person needle insertion view.

\begin{figure}[b]
    \centering
      \includegraphics[width=\linewidth]{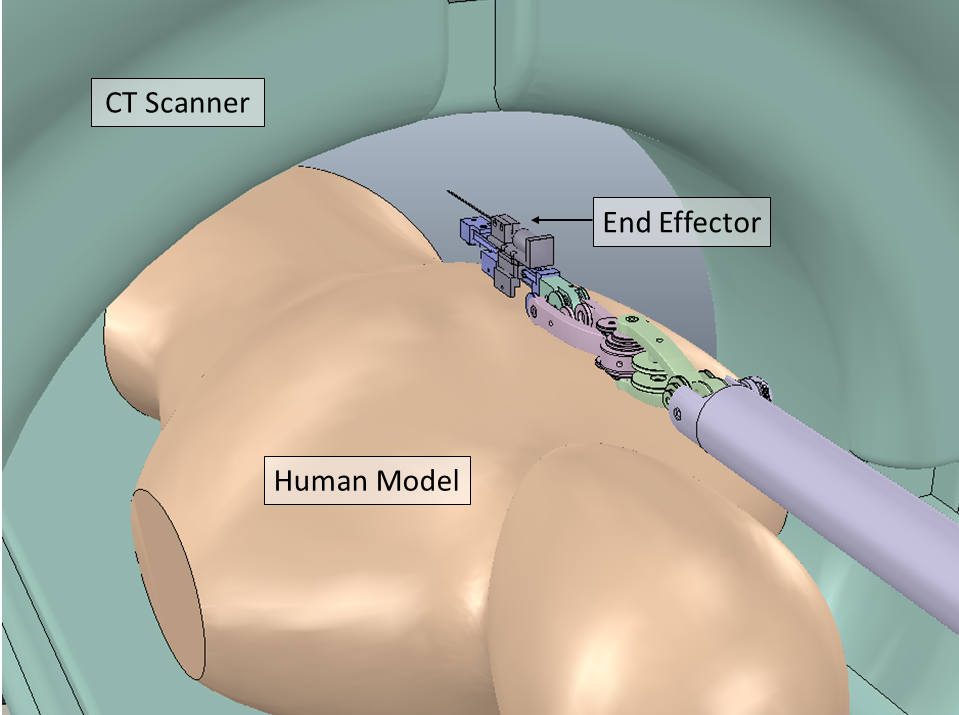}
    \caption{V-REP simulation environment shows the working space afforded by the serial linkage design that extends into the bore.}
    \label{fig:simulator}
\end{figure}

\section{Experiments and Results}
Experiments were conducted to show the performance and effectiveness of the system. Through simulation, the large reachable workspace on a human model in a CT bore is shown. A repeatability test is conducted to measure the precision of the system. Finally, the effectiveness of the system's teleoperational control is shown through a repeated needle targeting experiment and a in CT bore biopsy puncture of a phantom lung. The hardware test setup for the lab experiments are shown in Fig. \ref{fig:optitrack}. Due to limited parts on hand, the platform used in the hardware tests had shorter translational stage travels.


\subsection{Collision-free workspace evaluation}

\begin{figure}[t]
    \centering
    \vspace{2mm}
    \includegraphics[width=\linewidth,clip=true]{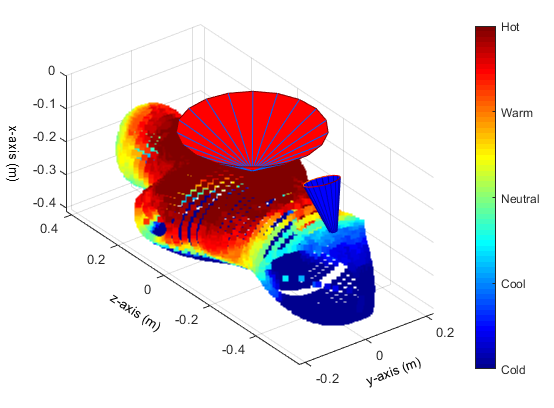}
    \caption{Heat map of the reachable workspace on an adult human in a CT bore. Hotter colors corresponds to higher dexterity. The cones show the angles of needle insertion possible for the example vertices.}
    \label{fig:workspace}
\end{figure}

V-REP was used to evaluate the collision-free workspace of the robot design. A virtual environment with a 65cm CT bore, a human adult dummy, and the robot are built as shown in Fig. \ref{fig:simulator}. Over one million ($N=1.77e6$) joint configurations for the first six joints were evaluated to determine the robot's reachable, collision-free work-space. The final insertion joint is excluded since it is used for needle insertion once reaching the target pose. The end-effector pose, and the overall robot's collision state (either self or environmental) are recorded for each point. In-collision points are removed. End-effector poses within a 5mm radius of each vertex of the Polygon File Format (.PLY file) robot model  of the dummy are binned and counted. The percentage of the populated cell are translated to the plots shown in Fig. \ref{fig:workspace}. Manual analysis of the lung region, which resides in the dark red region of Fig. \ref{fig:workspace} further highlights the dexterity of the robot. The two cones represent two extremes of the reachable workspace.

\subsection{Precision}
We tested system precision through a repeatability test. This illustrates the system's mechanical repeatability and tolerances. The robot repeated a sequence of 64 unique end-effector points, equally spaced through the robot's joint space, five times. At each point, the pose of the end-effector is recorded using an OptiTrack V120-Trio. The l2 deviation from the mean position and orientation (roll, pitch, yaw) for every configuration results in a standard deviation of 2.43mm and 2.95 degrees for position and orientation respectively. The contributed video illustrates this test.

\subsection{Teleoperated Accuracy}
\begin{figure}[t]
    \centering
    \vspace{2mm}
      \includegraphics[width=\linewidth]{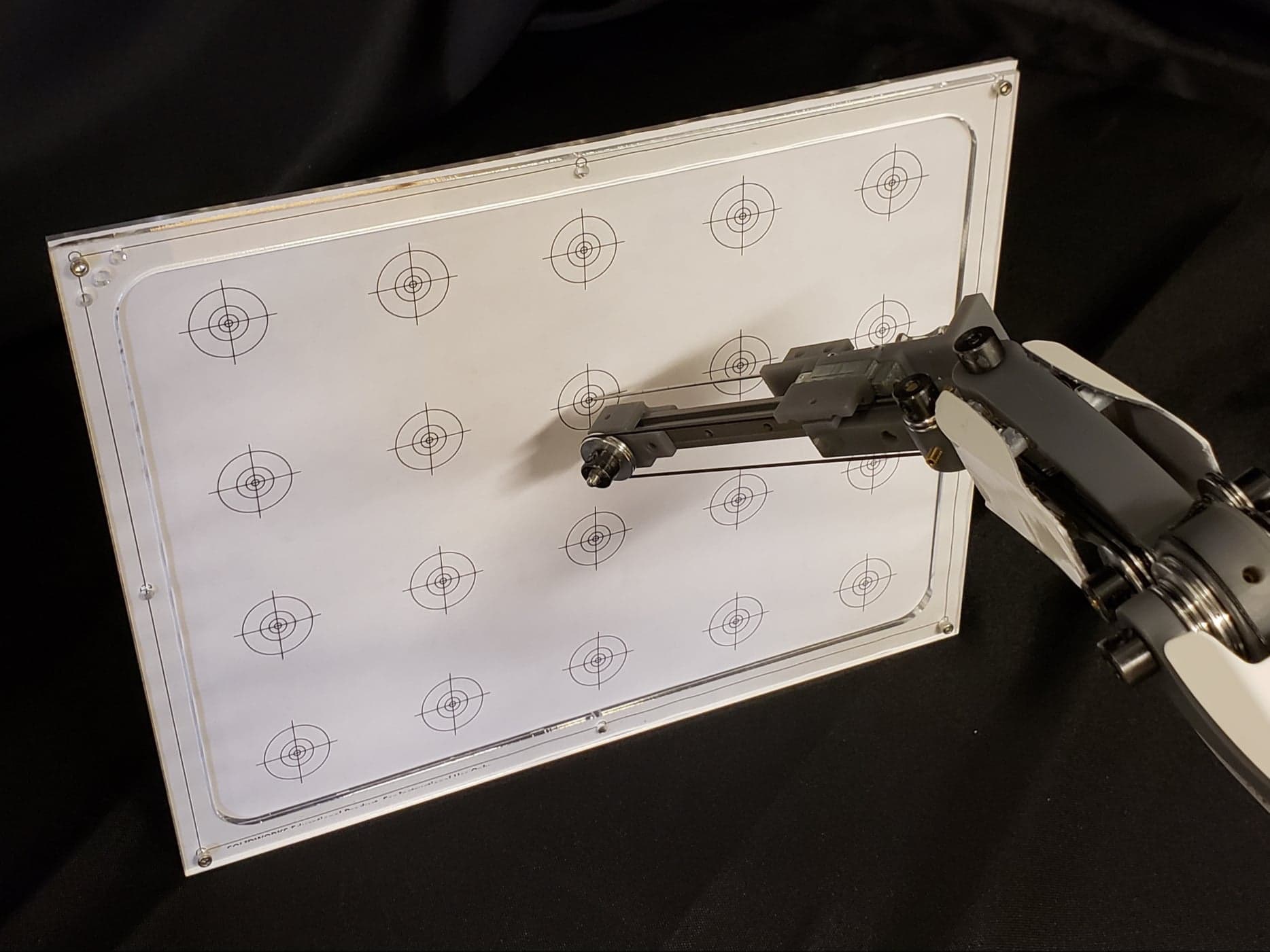}
    \caption{Test set up for the paper puncture target tests to verify teleoperational accuracy.}
    \label{fig:paperTarget}
\end{figure}

Using the needle and visual feedback from the  endoscopic camera, the operator controls the robot to puncture 16 bull's eyes on a four by four grid. The punctured paper target was scanned, and puncture distance from the bull's eye center is measured using the measuring tool within the GNU Image Manipulation Program \cite{gimp}. The target is shown in Fig. \ref{fig:paperTarget}. The mean positional error from the center of the bull's eye is 0.73mm with a standard deviation of 0.30mm.

\subsection{Biopsy task}

\begin{figure}[b]
    \centering
      \includegraphics[width=\linewidth]{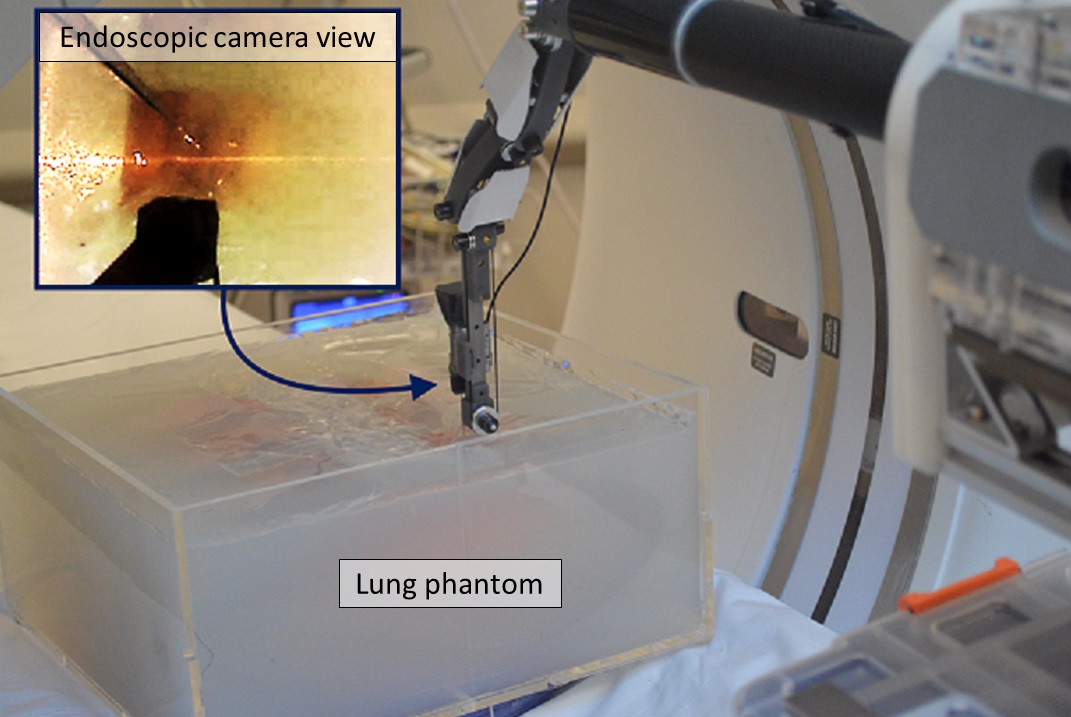}
    \caption{Photo showing a biopsy task where the robot was operated by a radiologist. The endoscopic camera view (inset) gives an additional perspective for the radiologist to ensure a precise needle insertion.}
    \label{fig:CT_Biopsy}
\end{figure}

\begin{figure*}
    \centering
    \vspace{2mm}
    \begin{subfigure}{0.49\linewidth}
          \centering
         \includegraphics[width=1\linewidth]{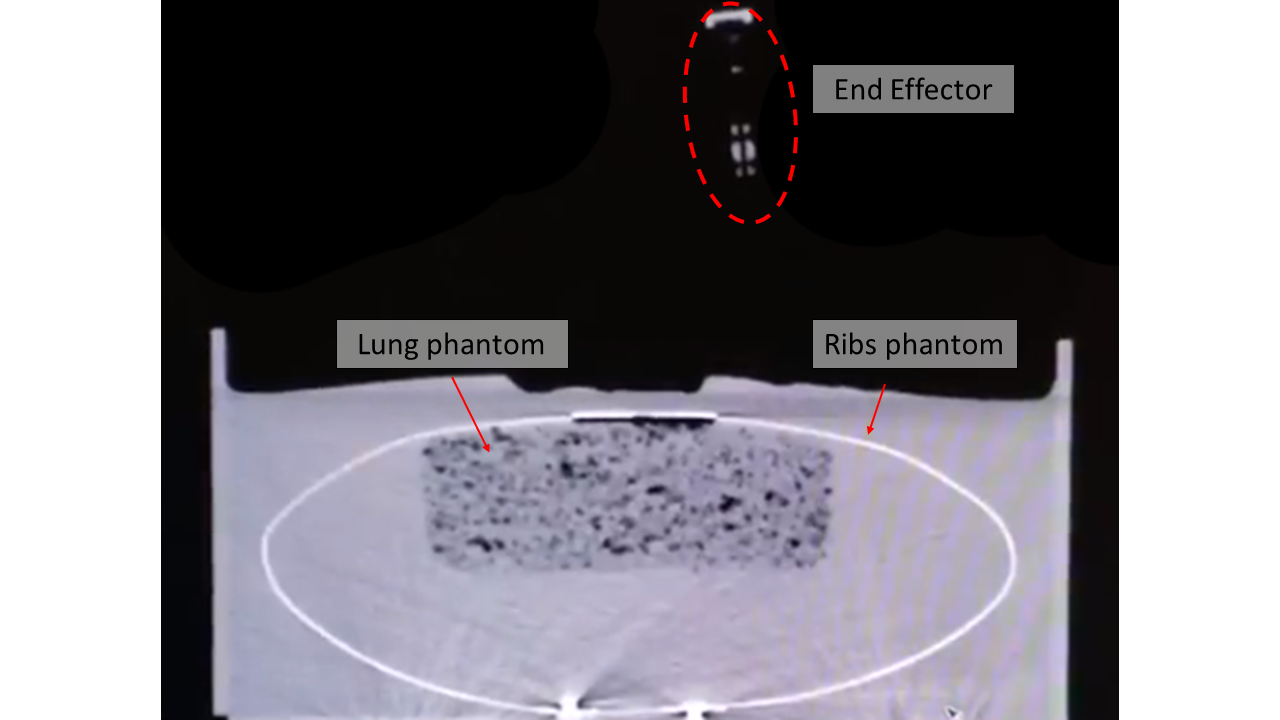}
          \caption{}
        \label{fig:CT_Biopsy_ribs}

    \end{subfigure}
    \hspace{1mm}
    \begin{subfigure}{0.49\linewidth}
        \centering
          \includegraphics[width=1\linewidth]{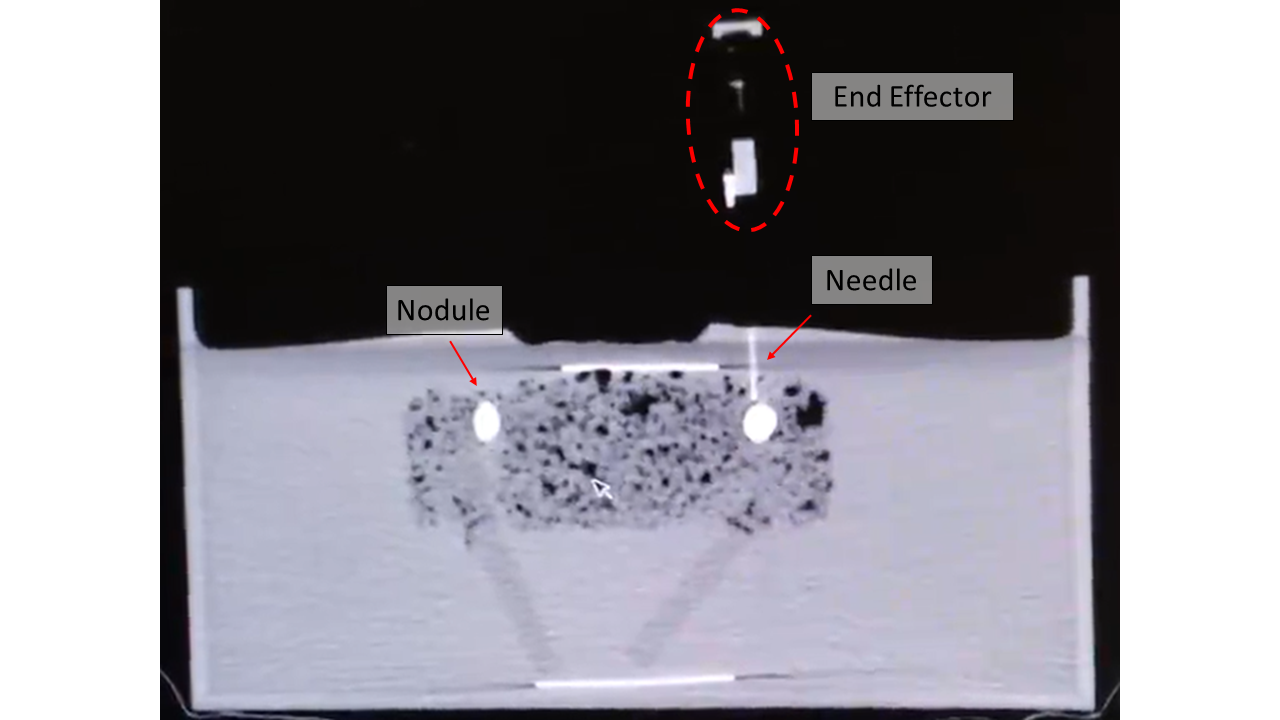}
          \caption{ }
          \label{fig:CT_Biopsy_puncture}
    \end{subfigure}
    \caption{CT scans from the teleoperated experiment. In (a), the phantom where the ribs and lungs are made from aluminum and sponge, respectively. In (b), the approach towards a needle biopsy of a nodule inside the phantom lung. (a) and (b) were taken at different $z$-depths.}
    \label{fig:CT_Biopsy_image}
\end{figure*}


The biopsy task was conducted inside a sliding stage CT scanner (GE 750HD) at the University of California San Diego's Thornton Hospital. Scan settings of 120 kVp, 40 mAs per slice, 0.4-second rotation time and a CTDI$_\text{vol}$ of 0.89 (similar to numbers reported in \cite{Tsai2009}) were used. Testing was performed on a custom lung phantom, based on the design presented in \cite{doi:10.1148/radiology.184.1.1609095}. The lung phantom consisted of a 7.25" $\times$ 5.25" $\times$ 2" synthetic foam sponge placed inside a sheet-aluminum ribcage. Two nodules were emulated with 1cm diameter clay balls and placed into the sponge. Targets were marked on top of the sponge to give rough position estimates for the nodules. Agar was cast in place over the ribcage and sponge. The entire lung phantom is held inside a 12" $\times$ 12.5" $\times$ 6" acrylic box.


An experienced radiologist teleoperated the robot using the 3DConnexion SpaceMouse to the target point (Fig. \ref{fig:CT_Biopsy}).For visual feedback, a combination of CT image data, an external webcam viewing the robot, the needle mounted endoscopic camera (Fig. \ref{fig:CT_Biopsy}), and the kinematic rendering provided in V-REP were used. Both targets were successfully reached on the first puncture attempt. The contributed video illustrates this test.

\subsection{CT interference Tests}
CT scans were performed to check for artifacts in the CT images due to metal in the end-effector. Sample images are shown in Fig. \ref{fig:CT_Biopsy_ribs} and \ref{fig:CT_Biopsy_puncture} where the needle was positioned vertically to show it on an transverse slice, showing no significant artifacts coming from the robotic system.

\section{Conclusion and Future work}
We present a teleoperated 7-DoF, low-profile and highly dexterous robotic needle placement platform for efficient and accurate needle biopsy of the lungs. Through the use of cable and belt drives, backlash is eliminated from the robot's transmission. Complete and detailed mechanical analysis, electrical design, user interface and systems design are available at \url{https://github.com/ucsdarclab/Open-Source-CT-Biopsy-Robot}. Simulation of the platform's collision-free workspace illustrated full access to the lungs from multiple orientations, repeatability with a standard deviation of less than 2.5mm and 3 degrees and minimal shadowing and artifacts in the CT scanner. Under teleoperation where the human closed-the-loop, mean position error of less than 0.75mm was achieved. Finally, the system is provided fully open-source to reduce the considerable development time for other researchers to engage in CT-guided robot biopsy research.  The system is designed with the flexibility and intra-bore compactness such that not only lung procedures can be performed, but most other image-guided interventional procedures may find it useful. 

In the future, we aim to develop the mechanical transmissions necessary to present a first-of-its kind, MRI-compatible serial link biopsy system. Furthermore, we aim to develop a unified user interface to present robot configuration and CT images in a single environment.


\balance
\bibliographystyle{ieeetr}
\bibliography{references}

\begin{thebibliography}{10}

\bibitem{cancerorg}
{American Cancer Society}, {\em Key Statistics for Lung Cancer}, 2019.
\newblock
  \url{https://www.cancer.org/cancer/non-small-cell-lung-cancer/about/key-statistics.html},
  Accessed: 2019-01-29.

\bibitem{Tian2017}
P.~Tian, Y.~Wang, L.~Li, Y.~Zhou, W.~Luo, and W.~Li, ``{CT-guided transthoracic
  core needle biopsy for small pulmonary lesions: Diagnostic performance and
  adequacy for molecular testing},'' {\em Journal of Thoracic Disease}, vol.~9,
  no.~2, pp.~333--343, 2017.

\bibitem{heerink2017complication}
W.~Heerink, G.~de~Bock, G.~de~Jonge, H.~Groen, R.~Vliegenthart, and M.~Oudkerk,
  ``Complication rates of ct-guided transthoracic lung biopsy: meta-analysis,''
  {\em European radiology}, vol.~27, no.~1, pp.~138--148, 2017.

\bibitem{kettenbach2015robotic}
J.~Kettenbach and G.~Kronreif, ``Robotic systems for percutaneous needle-guided
  interventions,'' {\em Minimally Invasive Therapy \& Allied Technologies},
  vol.~24, no.~1, pp.~45--53, 2015.

\bibitem{number46}
R.~Monfaredi, K.~Cleary, and K.~Sharma, ``Mri robots for needle-based
  interventions: Systems and technology,'' {\em Annals of Biomedical
  Engineering}, vol.~46, pp.~1479--1497, Oct 2018.

\bibitem{solomon2002robotically}
S.~B. Solomon, A.~Patriciu, M.~E. Bohlman, L.~R. Kavoussi, and D.~Stoianovici,
  ``Robotically driven interventions: a method of using ct fluoroscopy without
  radiation exposure to the physician,'' {\em Radiology}, vol.~225, no.~1,
  pp.~277--282, 2002.

\bibitem{anzidei2015preliminary}
M.~Anzidei, R.~Argir{\`o}, A.~Porfiri, F.~Boni, M.~Anile, F.~Zaccagna,
  D.~Vitolo, L.~Saba, A.~Napoli, A.~Leonardi, {\em et~al.}, ``Preliminary
  clinical experience with a dedicated interventional robotic system for
  ct-guided biopsies of lung lesions: a comparison with the conventional manual
  technique,'' {\em European radiology}, vol.~25, no.~5, pp.~1310--1316, 2015.

\bibitem{kuntz2016toward}
A.~Kuntz, P.~J. Swaney, A.~Mahoney, R.~H. Feins, Y.~Z. Lee, R.~W. III, and
  R.~Alterovitz, ``Toward transoral peripheral lung access: Steering
  bronchoscope-deployed needles through porcine lung tissue,'' in {\em Hamlyn
  Symposium on Medical Robotics}, pp.~9--10, 2016.

\bibitem{ben2018evaluation}
E.~Ben-David, M.~Shochat, I.~Roth, I.~Nissenbaum, J.~Sosna, and S.~N. Goldberg,
  ``Evaluation of a ct-guided robotic system for precise percutaneous needle
  insertion,'' {\em Journal of Vascular and Interventional Radiology}, 2018.

\bibitem{dou2017design}
H.~Dou, S.~Jiang, Z.~Yang, L.~Sun, X.~Ma, and B.~Huo, ``Design and validation
  of a ct-guided robotic system for lung cancer brachytherapy,'' {\em Medical
  physics}, vol.~44, no.~9, pp.~4828--4837, 2017.

\bibitem{moon2015development}
Y.~Moon, J.~B. Seo, and J.~Choi, ``Development of new end-effector for
  proof-of-concept of fully robotic multichannel biopsy,'' {\em IEEE/ASME
  Transactions on Mechatronics}, vol.~20, no.~6, pp.~2996--3008, 2015.

\bibitem{Zhou2011}
Y.~Zhou, K.~Thiruvalluvan, L.~Krzeminski, W.~H. Moore, Z.~Xu, and Z.~Liang,
  ``{An experimental system for robotic needle biopsy of lung nodules with
  respiratory motion},'' {\em 2011 IEEE International Conference on
  Mechatronics and Automation, ICMA 2011}, pp.~823--830, 2011.

\bibitem{Atashzar2013}
S.~F. Atashzar, I.~Khalaji, M.~Shahbazi, A.~Talasaz, R.~V. Patel, and M.~D.
  Naish, ``{Robot-assisted lung motion compensation during needle insertion},''
  {\em Proceedings - IEEE International Conference on Robotics and Automation},
  pp.~1682--1687, 2013.

\bibitem{Anzidei2015}
M.~Anzidei, R.~Argir{\`{o}}, A.~Porfiri, F.~Boni, M.~Anile, F.~Zaccagna,
  D.~Vitolo, L.~Saba, A.~Napoli, A.~Leonardi, F.~Longo, F.~Venuta, M.~Bezzi,
  and C.~Catalano, ``{Preliminary clinical experience with a dedicated
  interventional robotic system for CT-guided biopsies of lung lesions: a
  comparison with the conventional manual technique},'' {\em European
  Radiology}, vol.~25, no.~5, pp.~1310--1316, 2015.

\bibitem{walsh2007evaluation}
C.~J. Walsh, N.~Hanumara, A.~Slocum, R.~Gupta, and J.-A. Shepard, ``Evaluation
  of a patient-mounted, remote needle guidance and insertion system for
  ct-guided, percutaneous lung biopsies,'' in {\em ASME 2007 2nd Frontiers in
  Biomedical Devices Conference}, pp.~39--40, American Society of Mechanical
  Engineers, 2007.

\bibitem{Seitel2009}
A.~Seitel, C.~J. Walsh, N.~C. Hanumara, J.-A. Shepard, A.~H. Slocum, H.-P.
  Meinzer, R.~Gupta, and L.~Maier-Hein, ``{Development and evaluation of a new
  image-based user interface for robot-assisted needle placements with the
  Robopsy system},'' no.~March 2009, p.~72610X, 2009.

\bibitem{Hungr2016}
N.~Hungr, I.~Bricault, P.~Cinquin, and C.~Fouard, ``{Design and Validation of a
  CT-and MRI-Guided Robot for Percutaneous Needle Procedures},'' {\em IEEE
  Transactions on Robotics}, vol.~32, no.~4, pp.~973--987, 2016.

\bibitem{swaney2017toward}
P.~J. Swaney, A.~W. Mahoney, B.~I. Hartley, A.~A. Remirez, E.~Lamers, R.~H.
  Feins, R.~Alterovitz, and R.~J. Webster~III, ``Toward transoral peripheral
  lung access: Combining continuum robots and steerable needles,'' {\em Journal
  of medical robotics research}, vol.~2, no.~01, p.~1750001, 2017.

\bibitem{kratchman2011toward}
L.~B. Kratchman, M.~M. Rahman, J.~R. Saunders, P.~J. Swaney, and R.~J. Webster,
  ``Toward robotic needle steering in lung biopsy: a tendon-actuated
  approach,'' in {\em Medical Imaging 2011: Visualization, Image-Guided
  Procedures, and Modeling}, vol.~7964, p.~79641I, International Society for
  Optics and Photonics, 2011.

\bibitem{van2015design}
N.~J. van~de Berg, D.~J. van Gerwen, J.~Dankelman, and J.~J. van~den
  Dobbelsteen, ``Design choices in needle steering—a review,'' {\em IEEE/ASME
  Transactions on Mechatronics}, vol.~20, no.~5, pp.~2172--2183, 2015.

\bibitem{swaney2015tendons}
P.~J. Swaney, A.~W. Mahoney, A.~A. Remirez, E.~Lamers, B.~I. Hartley, R.~H.
  Feins, R.~Alterovitz, and R.~J. Webster, ``Tendons, concentric tubes, and a
  bevel tip: Three steerable robots in one transoral lung access system,'' in
  {\em Robotics and Automation (ICRA), 2015 IEEE International Conference on},
  pp.~5378--5383, IEEE, 2015.

\bibitem{reed2011robot}
K.~B. Reed, A.~Majewicz, V.~Kallem, R.~Alterovitz, K.~Goldberg, N.~J. Cowan,
  and A.~M. Okamura, ``Robot-assisted needle steering,'' {\em IEEE robotics \&
  automation magazine}, vol.~18, no.~4, pp.~35--46, 2011.

\bibitem{ion}
{Intuitive Surgical, Inc.}, {\em Ion}, 2019.
\newblock \url{https://www.intuitive.com/en/products-and-services/ion},
  Accessed: 2019-02-06.

\bibitem{auris}
{Auris Health, Inc.}, {\em Monarch Platform - Endoscopy Transformed}, 2019.
\newblock \url{https://www.aurishealth.com/monarch-platform}, Accessed:
  2019-02-06.

\bibitem{rojas2018robotic}
J.~R. Rojas-Solano, L.~Ugalde-Gamboa, and M.~Machuzak, ``Robotic bronchoscopy
  for diagnosis of suspected lung cancer: a feasibility study,'' {\em Journal
  of bronchology \& interventional pulmonology}, vol.~25, no.~3, p.~168, 2018.

\bibitem{fielding2017first}
D.~Fielding, F.~Bashirzadeh, J.~H. Son, M.~Todman, H.~Tan, A.~Chin, K.~Steinke,
  and M.~Windsor, ``First human use of a new robotic-assisted navigation system
  for small peripheral pulmonary nodules demonstrates good safety profile and
  high diagnostic yield,'' {\em Chest}, vol.~152, no.~4, p.~A858, 2017.

\bibitem{memoli2012meta}
J.~S.~W. Memoli, P.~J. Nietert, and G.~A. Silvestri, ``Meta-analysis of guided
  bronchoscopy for the evaluation of the pulmonary nodule,'' {\em Chest},
  vol.~142, no.~2, pp.~385--393, 2012.

\bibitem{gilbert2014novel}
C.~Gilbert, J.~Akulian, R.~Ortiz, H.~Lee, and L.~Yarmus, ``Novel bronchoscopic
  strategies for the diagnosis of peripheral lung lesions: present techniques
  and future directions,'' {\em Respirology}, vol.~19, no.~5, pp.~636--644,
  2014.

\bibitem{han2018diagnosis}
Y.~Han, H.~J. Kim, K.~A. Kong, S.~J. Kim, S.~H. Lee, Y.~J. Ryu, J.~H. Lee,
  Y.~Kim, S.~S. Shim, and J.~H. Chang, ``Diagnosis of small pulmonary lesions
  by transbronchial lung biopsy with radial endobronchial ultrasound and
  virtual bronchoscopic navigation versus ct-guided transthoracic needle
  biopsy: A systematic review and meta-analysis,'' {\em PloS one}, vol.~13,
  no.~1, p.~e0191590, 2018.

\bibitem{aerts2016transthoracic}
J.~G. Aerts, ``Transthoracic needle biopsies: It's more than just hitting the
  bull's-eye,'' {\em Clinical Cancer Research}, vol.~22, no.~2, pp.~273--274,
  2016.

\bibitem{number15}
Y.~Wang, W.~Li, X.~He, G.~Li, and L.~Xu, ``Computed tomography-guided core
  needle biopsy of lung lesions: Diagnostic yield and correlation between
  factors and complications,'' {\em Oncology Letters}, vol.~7, pp.~288--294,
  Jan 2014.

\bibitem{259616}
C.~J. Walsh, N.~C. Hanumara, A.~H. Slocum, J.-A. Shepard, and R.~Gupta, ``A
  patient-mounted, telerobotic tool for ct-guided percutaneous interventions,''
  {\em ASME Journal of Medical Devices}, vol.~2, no.~1, 2008.

\bibitem{VREP2013}
E.~Rohmer, S.~P.~N. Singh, and M.~Freese, ``V-rep: a versatile and scalable
  robot simulation framework,'' in {\em Proc. of The International Conference
  on Intelligent Robots and Systems (IROS)}, 2013.

\bibitem{gimp}
{GIMP}, {\em GNU Image Manipulation Program}, 2019.
\newblock \url{https://www.gimp.org/ }, Accessed: 2019-02-04.

\bibitem{Tsai2009}
I.~C. Tsai, W.~L. Tsai, M.~C. Chen, G.~C. Chang, W.~S. Tzeng, S.~W. Chan, and
  C.~C.~C. Chen, ``{CT-guided core biopsy of lung lesions: A primer},'' {\em
  American Journal of Roentgenology}, vol.~193, no.~5, pp.~1228--1235, 2009.

\bibitem{doi:10.1148/radiology.184.1.1609095}
W.~W. Scott and J.~E. Kuhlman, ``Phantom for use in lung biopsy training.,''
  {\em Radiology}, vol.~184, no.~1, pp.~286--287, 1992.
\newblock PMID: 1609095.

\end{thebibliography}

\end{document}